# Recursive Binary Neural Network Learning Model with 2.28b/Weight Storage Requirement

**Tianchan Guan[1,2], Xiaoyang Zeng[2], Mingoo Seok[1]**
[1]Electrical Engineering, Columbia University; [2]Microelectrical Engineering, Fudan University
{tg2569, ms4415}@columbia.edu; xyzeng@fudan edu

## Abstract

This paper presents a storage-efficient learning model titled Recursive Binary Neural Networks for sensing devices having a limited amount of on-chip data storage such as < 100's kilo-Bytes. The main idea of the proposed model is to recursively recycle data storage of synaptic weights (parameters) during training. This enables a device with a given storage constraint to train and instantiate a neural network classifier with a larger number of weights on a chip and with a less number of off-chip storage accesses. This enables higher classification accuracy, shorter training time, less energy dissipation, and less on-chip storage requirement. We verified the training model with deep neural network classifiers and the permutation-invariant MNIST benchmark. Our model uses only 2.28 bits/weight while for the same data storage constraint achieving ~1% lower classification error as compared to the conventional binary-weight learning model which yet has to use 8 to 16 bit storage per weight. To achieve the similar classification error, the conventional binary model requires ~4× more data storage for weights than the proposed model.

## 1. Introduction

Deep Neural Networks (DNN) have demonstrated the state-of-the-art results in a wide range of cognitive workloads such as computer version and speech recognition, achieving better-than-human performance for the tasks often considered too complex for machines [1-5]. The success of DNN has indeed motivated scientists and engineers to implement a DNN in mobile and embedded devices, often dubbed as Internet of *Smart* Things [6-7]. The recent works in this area however, mostly consider *off-device* learning [8-10], i.e., the learning of DNN is performed in the cloud data centers consisting of many CPU and GPU nodes and the post-learning weights are downloaded to mobile and embedded devices. This is because those devices have insufficient computing and storage capacity to train a DNN having millions of synapses each represented in a 32-bit floating point number.

*On-chip learning*, however, becomes increasingly important for the mobile and embedded devices for the following four reasons. First, such device often benefits to have the model that is custom-built for the device and its user and environment. This is because the model tends to be more accurate and effective if created with considering those factors. Second, if we cannot fit all the weights in on-chip storage we have to store them in off-chip storage such as FLASH and DRAM. Accessing off-chip storage, however, incurs 3 to 4 orders of magnitudes more energy and delay overhead than storage on a processor chip [15]. Uploading data onto cloud computers for training can consume even more power and substantially increase latency [11]. On-chip learning, therefore, is a very desirable approach for energy-efficiency and delay. Third, the training data from mobile and embedded sensing devices can contain security-sensitive information, e.g., personal health data from wearable medical devices. At the risk of being leaked, users typically do not want to upload such data onto cloud computers. Finally, in the era of Internet of Things (IoT), we anticipate a drastic increase in the number of deployed devices, which can proportionally increase the number of learning tasks to be done in the cloud. Coupled with the complexity of learning, even for powerful cloud computers, this can be computationally challenging tasks.

On-chip learning however entails various challenges, in algorithms, data, and systems [12-13]. The most eminent challenges with regards to hardware systems are high computation and data storage overhead of DNN system which can hardly be met by limited on-chip resources. Recently, Binary Neural Network (BNN) is proposed to drastically reduce computation complexity by using binary information of weights [16], activations [17], inputs [18], and their combinations. Although these works substantially save computational requirement, all these works still must maintain high-precision weights during training since the weight update is performed in a fine-grained manner. This limits on-chip integration of weights.

Scaling data storage requirement, however, is an urgent challenge to minimize off-chip storage and cloud computer access. For example, a wearable tracker such as FitBit, contains the ARM Cortex M3 processor. This processor has only 64 kilo-Byte (kB) on-chip data storage (data cache) [14]. Even if we use all of that storage only for weights, we can implement only 16,000 weights if each has to use 32 bits. Thus the device has to use the non-volatile FLASH memory chip with capacity up to tens of Mega-Byte (MB) to store synaptic weights. However, using it during learning would be prohibitive since it incurs 1,000-10,000X worse energy and delay to access them than the on-chip data storage of the processor [15].

Our goal is, therefore, to train a neural network by efficient using a limited amount of data storage on a processor chip. Toward this goal, we propose a new learning model, *Recursive Binary Neural Network (RBNN)*. This model is based on the process of training of a neural network, weight binarization, recycling storage of non-sign-bit portion of weights to add more weights to enlarge the neural network for performance improvement. We recursively perform this process until either accuracy stop improving or we use up all the storage on a chip.

We verified the proposed RBNN model on a multi-layer perceptron (MLP)-like classifier and the MNIST benchmark. We considered typical storage constraints of embedded sensing devices in the order of hundreds of kB. The experiment confirms that the proposed model (i) demonstrates ~1% classification accuracy improvement over the conventional BNN learning model specifically following [16] for the same storage constraints or (ii) scale on-chip data storage requirement by 4X for the same classification test error (~2.5%), marking the storage requirement of 2.28 bits/weight. The conventional BNN model ([16] but also [17,18]) exhibits a significantly larger storage requirement of 8 to 16 bits/weight..

The remainder of the paper is as follow. In Sec. 2 we will introduce the existing works on . In Sec. 3 we will describe our proposed model. Sec. 4 will present the experimental results and comparisons to the conventional model. Finally, in Sec. 5, we will conclude the paper.

## 2. Related Work

While DNN achieves the state-of-the-art classification accuracy in a range of cognitive tasks, DNN is also well-known for its high computational complexity and storage requirement. The high storage requirement of DNN has motivated active researches to mitigate it both in learning and inference processes. Particularly many of the researches focus on synaptic weights as the high storage requirement often lies in and around them conventionally represented in 32-bit floating-point numbers.

### 2.1. Synaptic weight compression

Compressing weights is one of the approaches to reduce storage requirement in deployment. Several researches show that the eights of DNN contain a large amount of redundancy, which can be exploited by compression. Well-known compression algorithms such as hash function [19] and Huffman code [20] have been applied to compress weights in DNN, achieving a large amount of storage savings as high as 49× [20]. However, these approaches must compress weights only after finishing training, thus *cannot scale storage requirement for learning*. Furthermore, it requires a decompression process of weights before performing inference operation, incurring a considerable amount of additional computing complexity.

### 2.2. Synaptic weights precision reduction

Another approach is to reduce the precision of synaptic weights. Several studies have demonstrated lowering the precision of weights (i.e., quantization) has a tolerable impact on the performance of DNN [21,22]. Some studies even showed that the noise introduced by the quantization process can help improving performance [22]. In Ref. [21], the authors trained a DNN having 16-bit fixed-point weights with the proposed stochastic rounding technique, and demonstrated little to no degradation in classification accuracy. In Ref. [22], the authors proposed the dynamic fixed-point representation (i.e., dynamically changing the position of decimal point over computation sequences) to further reduce the precision requirement down to 10 bits per synapse. These techniques help to reduce storage requirement and replace complex floating-point arithmetic with simple fixed-point one.

### 2.3. Binary Neural Network

Recent works proposed to use binary information of weights [16], activations [17,18], and even inputs [18] for some parts of learning and post-learning operations of neural networks. This replaces a good number of real-number multiplications with sign-inversion, shift, and XNOR operations, largely reducing computational complexity. However, all of these BNN approaches must store *high-precision weights* and cannot scale storage requirement of weights during *learning* because it is needed to tune weights in a fine-grained manner.

## 3. RBNN Model

### 3.1. Key idea

Intuitively, our RBNN model is inspired by the difference between precision requirement of weights for BNN training and inference. To shed light with this difference, Table I shows which parts of weights are used in each step of training. The conventional BNN works [16-18] can use only sign bits of weights during multiply-and-accumulate (MAC) in both forward and back propagations. However the weight update is done in high-precision. This requires to store all bits of weights during learning i.e., no scaling in storage savings.

**Table I Comparisons of weight information usage in BNNs and RBNN**

| Steps | BNN [16-18] | Proposed RBNN |
|---|---|---|
| **MAC in forward prop.** | Sign bits of weights | Sign bits of weights |
| **MAC in back prop.** | Sign bits of weights | Sign bits of weights |
| **Weight update** | All bits of weights | All bits of weights |
| **Recursive recycling** | N/A | Keep sign bits and recycle storages of the other bits for more plastic weights |

However, it has been known that once trained we can use only sign bits of weights to perform inference at a reasonably small accuracy penalty [16-18]. This vast different requirement of weight precision between learning and post-learning inspires us to create our RBNN model. As shown in the third column in Table I, like BNN [16-18], we also use sign bits for MAC operations to reduce computational complexity. The main difference is that after keeping only the sign bits after learning (called binarization), we recycle storages that used to store non-sign bits of weights and add more weights to the neural network. We perform this process recursively, which makes the neural networks larger and more accurate but using the same amount of storage.

Figure 1 depicts the process of our proposed RBNN learning model with an example fully-connected neural network. In the beginning the neural network has one input, two hidden, and one output neurons with four weights each of which has n bits. We first train this $1 \times 2 \times 1$ network using the conventional back-propagation algorithm. After that, as conventional BNN training method does, we binarize all the weights, i.e., discarding all bits except the sign bit, resulting in a $1 \times 2 \times 1$ trained network with binary weights (trained BNN). Then we continue the second iteration of training (the middle of Fig. 1): specifically, we recycle the storage that used to store the *n-1* non-sign bits of synapses in previous network. Using this storage, we add four weights each of which is now (n-1) bits to the just trained network, expanding the network to $1 \times 4 \times 1$. This network has four weights that are trained (non-plastic, marked as solid lines) and four weights that can be trained (plastic, marked as dash lines), and we train only those four plastic weights using back-propagation algorithm.

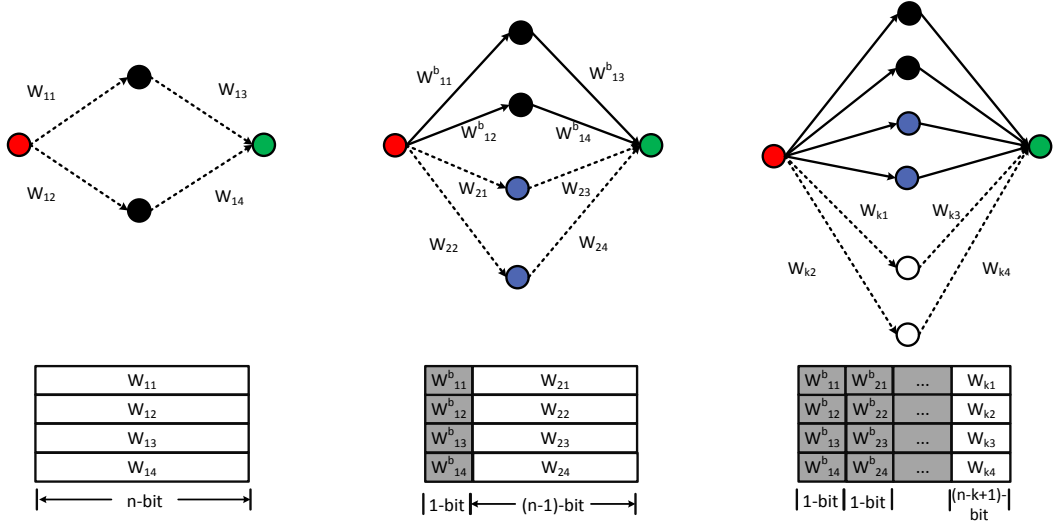

Figure 1: RBNN learning model with an example neural network. The recursive operation increases the number of weights in the neural network (top) while using the same amount of storage for weights (bottom)

We repeat the same process of binarization and recycling, and after the k-times iterations the network becomes $1 \times 2 \cdot k \times 1$ with $4 \cdot k$ binary weights, k times more weights than the first network. However, the storage for weights we need to use did not grow over the iterations, scaling the storage requirement per weight to n/k (=4·n/4·k), again k times better than the first network. Thus the proposed RBNN either can achieve better classification accuracy - enabled by the more number of weights - or reduce weight storage requirement to achieve the same performance level.

### 3.2. Model details

Figure 2 depicts the details of the RBNN model. The left part of Figure depicts the flow chart of RBNN model, while left part explains the function of each stage. In the beginning of the training procedure, conventional BNN training algorithm *BNN_Training* is used to train a BNN. After training, we got a trained BNN with binary synaptic weights. As a result, the synaptic bit-width is reduced by 1. And then we use the rest synaptic bits are used as weights of *incremental BNN*. After training the *incremental BNN* with algorithm *Incremental_BNN_Training*, the performance of the *enlarged BNN* is tested. If the performance doesn't stop improving and there are still available synaptic bits after weight binarzation, the rest synaptic bits will be reused to further enlarge current *trained BNN*.

The method *Incremental_BNN_Training* is designed to train the *incremental BNN* to improve performance of previously *trained BNN*. To meet this goal, the conventional BNN training method is adjusted as follows:

**Feedforward:** The outputs of enlarged BNN computed through adding outputs of both *trained BNN* and *incremental BNN*. As shown in Figure1, there're no connections between trained BNN and incremental BNN. So the outputs of each hidden layer in the two networks are calculated separately.

**Backpropagation & Parameter update:** The output layer's activation gradient only back-propagates to incremental BNN. And only synaptic weights in incremental BNN are updated.

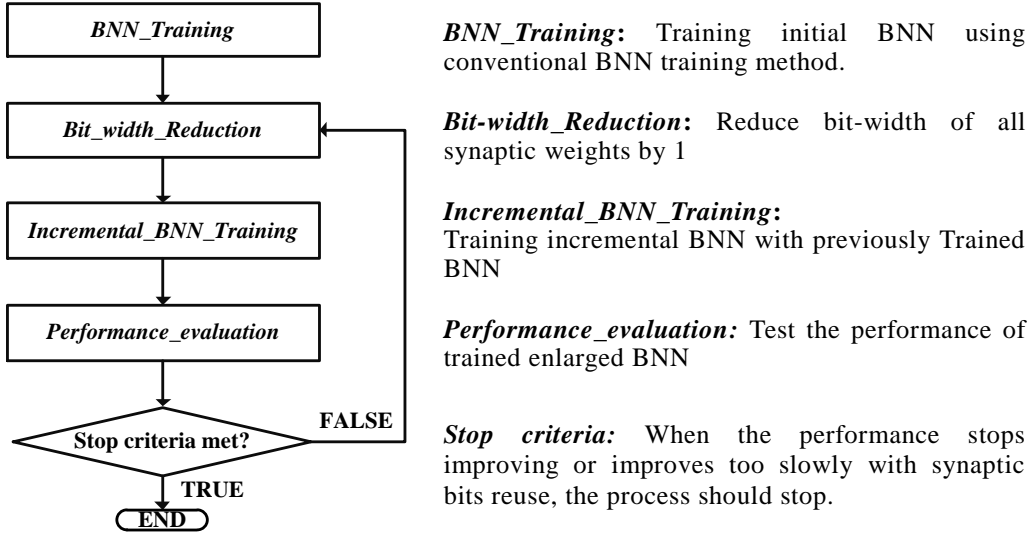

*BNN_Training*: Training initial BNN using conventional BNN training method.

*Bit-width_Reduction*: Reduce bit-width of all synaptic weights by 1

*Incremental_BNN_Training*:
Training incremental BNN with previously Trained BNN

*Performance_evaluation:* Test the performance of trained enlarged BNN

*Stop criteria:* When the performance stops improving or improves too slowly with synaptic bits reuse, the process should stop.

Figure 2: Detail of Training Method for RBNN Model

To sum up, the trained BNN is only used to calculate output of the enlarged network in feedforward stage. The parameters of trained BNN won't change when training i*ncremental BNN*. This ensures convergence of training of the *enlarged BNN*. The weights in *trained BNN* are binary, so it can only capture small partition of the weight updates because most of them have very small amount. Therefore, if we update *trained BNN*, it is very likely that the algorithm will not converge to a minimum of the cost function. On the other hand, the weights of *incremental BNN* have relatively high precision during training and thus can capture most of the weights update.

Only updating incremental BNN can reduce the cost function of the enlarged BNN is because there's no connections between *incremental BNN* and *trained BNN*. Therefore, when calculating output of *enlarged BNN*, output of *trained BNN* can be considered as a constant added to output of *incremental BNN*. As such, during back-propagation, *trained BNN* outputs have no impact on gradient of *incremental BNN* because derivation of constant is zero.

Similar to the conventional BNN training algorithm [16], binary weights are used in both feed forward and back propagation in *Incremental_BNN_Training*, to reduce computational overhead. Since weights in *trained BNN* are binary, the multiplication related to weights can all be simplified as shift.

---

**Algorithm** 1 *Incremental_BNN_Training*. $C$ is the cost function for mini-batch, $\eta$ the learning rate and $L$ the number of layers. The function $Binarize()$ specifies how to binarize the weights. $Act\_hid()$ and $Act\_out()$ are activation function of hidden layers and output layer, respectively.

**Require**: a minibatch of inputs and targets ($a_0$, $a^*$), previous weights of incremental BNN $W(I)$, weights of trained BNN $W(T)$,
**Ensure**: updated weights of incremental BNN $W(I)^{t+1}$
{1. Forward Propagation}
{1.1 Computing outputs of hidden layers in trained BNN and incremental BNN}
**for** $k$ = 1 to $L$-1 **do**

$$a(T)_k = Act\_hid(W(T)_k \cdot a(T)_{k-1})$$
$$W(I)_k^b \leftarrow Binarize(W(I)_k^b)$$
$$a(I)_k = Act\_hid(W(I)_k^b \cdot a(I)_{k-1})$$
**end for**
{1.2 Computing outputs of enlarged BNN}
$$a_L = Act\_out(W(T)_L \cdot a(T)_{L-1} + W(I)_L \cdot a(I)_{L-1})$$
{2. Back propagation}
{Please note that only gradients of **incremental BNN** are computed.}
Compute $g_{a_L} = \frac{\partial C}{\partial a_L}$ knowing $a_L$ and $a^*$
**for** $k = L$ to 1 **do**
$$g_{W(I)_k^b} \leftarrow \left(g_{a(I)_k} \cdot W(I)_k^b\right) \circ a'(I)_{k-1}$$
$$g_{W(I)_k^b} \leftarrow g_{a(I)_k}^T \cdot a(I)_{k-1}$$
**end for**
{3 Parameter Update}
{Please note that only weights of **incremental BNN** are updated.}
**for** $k = L$ to 1 **do**
$$W(I)_k^{t+1} \leftarrow W(I)_k^t + \eta \cdot g_{W(I)_k^b}$$
**end for**

---

## 4. Experiment Setup

### 4.1. Permutation-invariant MNIST benchmark

We used the permutation-invariant MNIST to test the performance of the proposed RBNN model. We use the original training set of 60,000 images and the original test set of 10,000 28-by-28 pixel gray-scale images. The training and testing data $x$ is normalized to $x_{norm}$ as:

$$x_{norm} = \frac{x - mean(x)}{255} \times 2 \qquad (7)$$

$x_{norm}$ is in the interval [-1, 1] and exhibits zero mean. Following the typical practices, we use the last 10,000 images of the training set as a validation set for early stopping and model selection. As we use the permutation-invariant MINST, i.e., ignoring the 2-dimentonal image structure of the image, we did not consider convolutional computation. We also did not consider data augmentation, pre-processing, and unsupervised pre-training during our experiment.

### 4.2. Neural network configuration and data format

We considered the storage constraints of mainly hundreds of kB based on the typical embedded system designs [23]. We considered a feed-forward fully-connected neural network with a single hidden layer. We considered several different numbers of neuron units in the hidden layer ranging from 200 to 800. The numbers of the input and output units are 784 and 10, respectively. We used the *tanh_opt()* for the activation function of the hidden layer and the *softmax()* for that of the output layer. We use the classical Stochastic Gradient Descent (SGD) algorithm for cross-entropy minimization without momentum. We use a small size of batch (1,000) and a single static learning rate of 0.25. We did not use other advanced techniques such as dropout, Maxout, ADAM, and etc. for both the proposed and the baseline learning models. We recorded the best training and test errors associated with the best validation error after up to 1,000 epochs.

We used the fixed-point arithmetic for all the computation and data load and access. The synaptic weight data load and access was performed strictly at the precision of the plastic synaptic bit-width associated with the current recursive iteration $BW_p$, e.g., N bits for the first iteration, N-1 bits for the second iteration, et al. The intermediate computation, such as gradient calculation, may use higher-precision fixed-point arithmetic to maintain sufficient precision. The translation from wide fixed-point number to narrow fixed-point and binary number is performed with simple decimation, i.e., removing less significant bits, without using advanced techniques such as stochastic rounding. We saturated values in the event of overflow in weight update, i.e., the value was set to the largest or smallest value which the currently-used fixed-point format can represent.

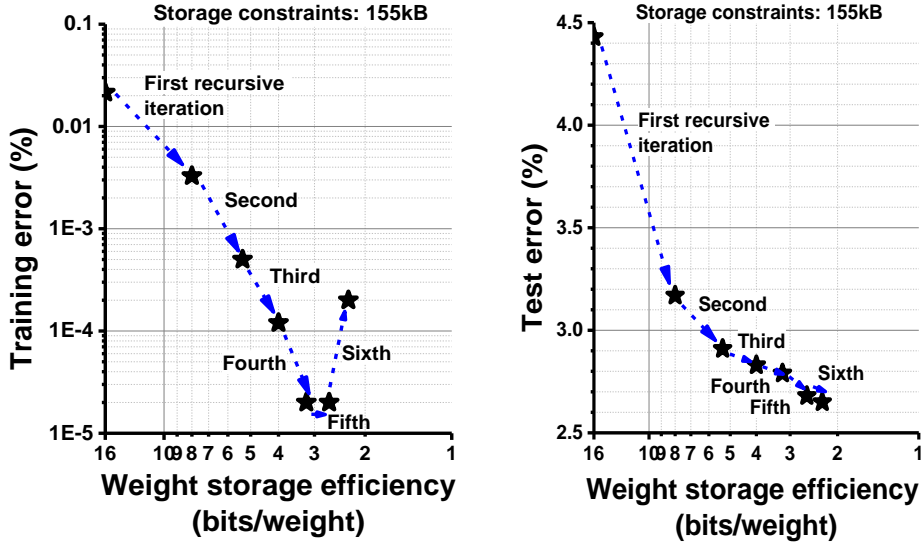

Figure 3: (left) Training error and (right) testing error across recursive iterations in the proposed RBNN model. The storage constraint is 155 kB.

## 5. Results and Discussion

### 5.1. Accuracy improvement

Figure 3 depicts the classification errors of the proposed learning model across six recursive iterations. The initial bit-width of weights is 16 bits. In each series of data points in Figure 3, the leftmost point is the initial neural network, i.e., with 100 hidden units and 79,400 synapses (= 784·100+100·10). At this point, the synaptic bit efficiency, defined as *the ratio of total storage bits to the number of synapses trained with the bits*, is 16 bits/synapse. Note that this point is equivalent to the network trained with the conventional BNN model specifically following [16]. In the second leftmost data point in the series is the neural network after the first recursive iteration, i.e., the network with total 200 hidden units and 158,800 (=2·79,400) synapses, achieving significant reduction in error. The weight storage requirement also improves to ~8 bits/synapse. Finally, after six recursive iterations, the neural network implements total 700 hidden neurons and 555,800 synapses with only 155 kB data storage, marking the storage requirement of 2.28b/synapse and the test error of ~2.6%.

### 5.2. Storage savings

We experimented our proposed and conventional BNN learning model [16] across different combinations of hidden neuron counts and bits per synapse. For the conventional model, we considered from 800 hidden neurons and 16-b synapses to 100 hidden neurons and 12-b synapses, which correspond to 1.2 MB to 116 KB data storage requirement for synaptic weights, respectively. For the proposed model, we considered from 200 *initial* hidden neurons and initially 16 bit synapses to 100 initial hidden neurons and initially 12 bit synapses, and performed recursive binarizations for each network. Our proposed model requires 116kB to 310 kB data storage for synapses. Figure 4 shows the results of this experiment: the proposed model can achieve ~1% lower test error than the conventional model using the similar amount of storage. In addition, the conventional mode requires 3× more data storage to achieve the similar test error with the proposed RBNN model.

Table II shows the detail comparisons of the six neural networks, three from the proposed RBNN model ($R_1$, $R_2$, $R_3$) and three from the conventional BNN model ($B_1$, $B_2$, $B_3$) [16]. $R_1$ and $B_1$ achieve the similar test error, but $R_1$ outperforms $B_1$ in the storage requirement by $3\times$. The downside of $R_1$ is the increase in computations during training: it requires more shift and add operations, but the same number of multiplications (which is much more complex than add and shift), as compared to the B1. The computation complexity for inference operations are the same. Similarly, $R_3$ requires $4\times$ less storage than $B_2$.

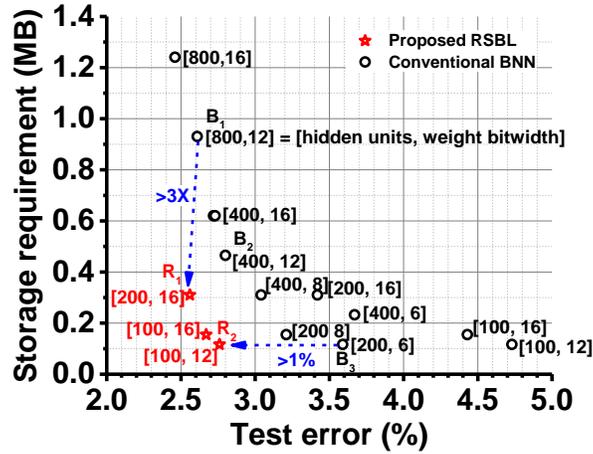

Figure 4: The storage requirement and test error trade-offs achieved by the proposed RBNN model and the conventional BNN model. The proposed model achieves >3× storage requirement savings for the same test error and >1% lower error for the same storage requirement.

Table II. Detail comparisons of RBNNs and BNNs

|  | $R_1$ | $R_2$ | $R_3$ | $B_1$ | $B_2$ | $B_3$ |
|---|---|---|---|---|---|---|
| *Initial hidden neurons* | 200 | 100 | 100 | 800 | 400 | 200 |
| *Final hidden neurons* | 800 | 700 | 400 | 800 | 400 | 200 |
| *Final synapses* | 635,200 | 555,800 | 317,600 | 635,200 | 317,600 | 155,600 |
| *Initial weight bitwidth* | 16 | 16 | 12 | 12 | 12 | 6 |
| *Storage req (b/weight)* | 4 | 2.28 | 3 | 12 | 12 | 6 |
| *Test error (%)* | 2.56 | 2.65 | 2.76 | 2.61 | 2.80 | 3.60 |
| *Comp., learning Shift / Multiply / Add* | 2,699,600 635,200 2,699,600 | 3,970,000 555,800 3,970,000 | 1,349,800 317,600 1,349,800 | 1,270,400 635,200 1,270,400 | 635,200 317,600 635,200 | 317,600 158,800 317,600 |
| *Comp., inference Shift, Add* | 635,200 635,200 | 555,800 555,800 | 317,600 317,600 | 635,200 635,200 | 317,600 317,600 | 158,800 158,800 |
| *Storage requirement* | 310kB | 155kB | 116kB | 930kB | 465kB | 114kB |

## 6. Conclusion and Future Work

This paper presents a learning model regarding local on-device learning in sensing devices with limited data storage. The proposed RBNN model builds upon the recent binary neural network models and extends them by recycling data storage that would have been wasted, to add and train more synapses to a neural network classifier. We verified the proposed model with the neural network classifier and the permutation-invariant MNIST benchmark under the typical embedded system storage constraints. The results show that the proposed model achieves 2.28b/weight storage requirement while achieving ~1% better classification error as compared to the conventional binary-weight learning model for the same storage constraint.

Our proposed model also achieves 3× less data storage than the conventional model for the same classification error.

We expect the future work that extends the application of the learning model to other neural network topologies and datasets. We also expect to apply the RBNN model to the ensembles of neural networks [24-26], the mixture of experts [27-29], and the incremental learning [27,30]. This is because the proposed learning model virtually enables to create and train additional synapses from the just trained synapses via recursive binarization and recycling.